%% file: main.tex
\documentclass[10pt,twocolumn,letterpaper]{article}

\usepackage{cvpr}              

\input{preamble}

\definecolor{cvprblue}{rgb}{0.21,0.49,0.74}
\usepackage[pagebackref,breaklinks,colorlinks,allcolors=cvprblue]{hyperref}

\def\paperID{jzAt79drAV} 
\def\confName{CVPR}
\def\confYear{2026}

\title{Metadata-Aware Multi-Prompt Reasoning for Zero-Shot Accident Understanding}

\author{Tarandeep Singh\thanks{Equal contribution.} \quad Soumyanetra Pal\footnotemark[1] \quad Soham Biswas \quad Nishanth Chandran\\
Netradyne\\
{\tt\small \{firstname.lastname\}@netradyne.com}
}

\begin{document}
\maketitle
\input{sec/0_abstract}    
\input{sec/1_intro}
\input{sec/2_related}
\input{sec/3_method}
\input{sec/4_experiments}
\input{sec/5_conclusion}
{
    \small
    \bibliographystyle{ieeenat_fullname}
    \bibliography{main}
}

\input{sec/X_suppl}

\end{document}

%% file: preamble.tex
\usepackage{booktabs}
\usepackage[most]{tcolorbox}
\usepackage[margin=0.87in]{geometry}



%% file: sec/0_abstract.tex
\begin{abstract}
In this paper, we address the problem of zero-shot understanding
of accidents from surveillance videos by identifying \emph{when}
an impact event occurs, \emph{what} type of impact it is, and
\emph{where} in the frame it occurs using natural language.
We propose a three-stage pipeline that decomposes the accident
understanding into when, what, and where. The first stage extracts
a short temporal window around the impact using vision-language
similarity. In the second stage, we perform metadata-driven
multi-prompt reasoning with five complementary views (baseline,
motion, geometry, contrast, and tiebreaker) and resolve disagreement
via an entropy-gated pairwise adjudicator. Finally, we localise
the impact of an open-vocabulary detector queried on the
predicted accident type and scene layout, and aggregate detections
across keyframes using a score-weighted centroid. Our pipeline
achieves a substantial improvement in the harmonic-mean score over a
centre-of-frame baseline on the zero-shot ACCIDENT @ CVPR benchmark.
We show that decomposing zero-shot video understanding into temporal
localisation, semantic classification, and spatial grounding enable
more reliable reasoning with vision-language models than direct
prompting alone.
\end{abstract}

%% file: sec/1_intro.tex
\section{Introduction}
\label{sec:intro}

Accident understanding in surveillance video is an important problem
for emergency response, insurance assessment, fleet safety, and
autonomous driving. In these settings, it is not enough to determine
whether an accident has occurred. It must also identify
the time of the incident, recognise the accident category, and localise
the physical point of impact.

Although this may appear straightforward for short video clips, it
remains difficult in real-world CCTV footage. Accident categories can be
visually similar, and the decisive impact often occurs within
only a few frames. As a result, models may attend to contextual cues or
prominent objects rather than the actual interaction between road
agents. These challenges are amplified in zero-shot scenarios such as
the ACCIDENT@CVPR 2026 challenge~\cite{picek2026accident}, where models
must generalize to unconstrained real surveillance
videos without labelled real-world training examples.

Vision-language models (VLMs) offer a natural way to approach this
problem because they can interpret visual evidence through textual
queries. However, formulating accident analysis as a single end-to-end query
requires the model to make multiple complex decisions at once, and the
prediction is often unstable and prone to shortcuts.
Prior work has shown that decomposing visual understanding into
targeted questions~\cite{antol2015vqa,malla2022drama} and aggregating
over multiple predictions~\cite{wei2022cot,wang2022selfconsistency}
improves robustness.

In this work, we propose a three-stage zero-shot pipeline organised
around the questions \textit{when}, \textit{what}, and \textit{where}
(Fig.~\ref{fig:pipeline}). First, the \textit{when} stage selects a
compact temporal window likely to contain the collision by combining
vision-language similarity with motion-based cues. Second, the
\textit{what} stage predicts the accident type using several structured
prompts that emphasise complementary aspects of the scene; inconsistent
outputs are resolved using a lightweight adjudication step. Finally,
the \textit{where} stage localises the impact region with an
open-vocabulary detector conditioned on the predicted accident type and
scene context, followed by aggregation across frames.

The proposed system only uses open-weight models, requires no
fine-tuning, and runs on a single 24\, GB GPU. Experiments show that
explicitly decomposing accident understanding into temporal detection,
type recognition and spatial grounding yield more reliable zero-shot
performance than asking a model to solve the full task in one step.

Our main contributions are:
\begin{itemize}
  \item A three-stage \textit{when}/\textit{what}/\textit{where}
    framework for zero-shot accident understanding from CCTV
    video, with each stage independently upgradable.
  \item A five-prompt classification scheme with entropy-gated
    pairwise adjudication that resolves ambiguous votes without
    labelled calibration data.
  \item A type and scene conditioned localization strategy that improves harmonic-mean score, highlighting a foreground bias in existing pointing models.
\end{itemize}

%% file: sec/2_related.tex
\section{Related Work}
\label{sec:related}

\begin{figure*}[ht]
\centering
\includegraphics[width=0.9\textwidth]{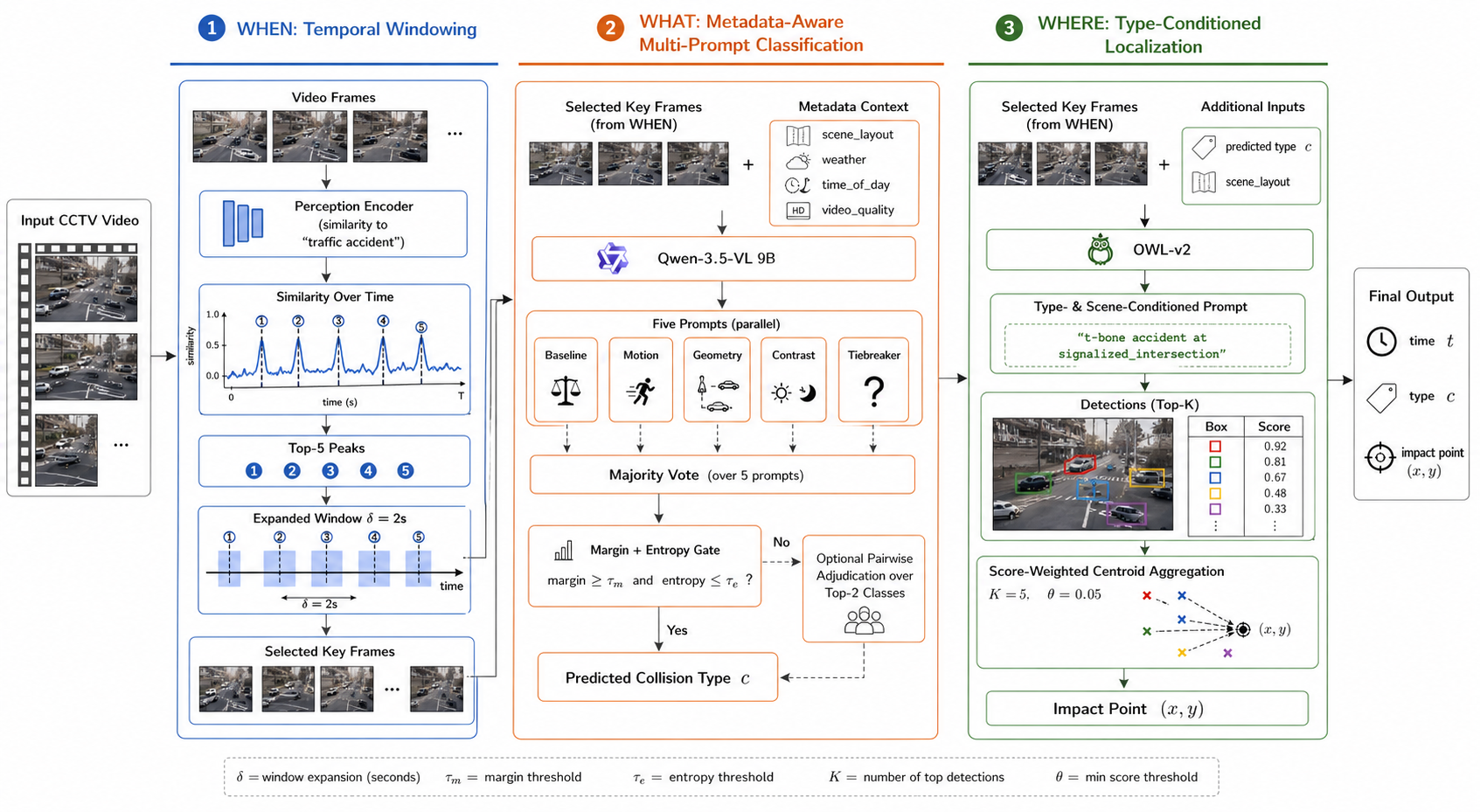}
\caption{Three-stage When–What–Where pipeline for accident understanding: temporal localization, metadata-conditioned multi-perspective classification, and type-guided spatial impact localization for accurate, interpretable predictions.}
\label{fig:pipeline}
\end{figure*}

\paragraph{Accident detection and anticipation.}
Traffic accident analysis has traditionally focused on detecting
whether an accident occurred or anticipating whether one is about to
happen. Early work such as DSAR-NN~\cite{chan2016dsarnn} used
spatial attention and recurrent modeling for accident anticipation
from dashcam videos. CADP~\cite{shah2018cadp}, DoTA~\cite{yao2020dota},
and TAD~\cite{xu2024tad} expanded this direction with datasets for
traffic anomaly and accident detection, while Fang
\etal~\cite{fang2023survey} survey the broader field. These works
primarily emphasize event detection and temporal localization; accident
type is often treated as an auxiliary label. In contrast, we explicitly
separate the three outputs required by ACCIDENT@CVPR 2026~\cite{picek2026accident}: accident
time, collision type, and impact location.

\paragraph{Driving scene understanding with language.}
Language has been used to improve interpretability in driving videos,
including explanations of ego-vehicle behavior in BDD-X~\cite{kim2018bddx},
object-level action explanations in BDD-OIA~\cite{xu2020bddoia}, and
causal reasoning over driver behavior in HAD/HDD~\cite{kim2019had,ramanishka2018hdd}.
Most relevant to our formulation is DRAMA~\cite{malla2022drama}, which
uses structured questions about risky objects --- what they are, where
they are, and why they matter --- to improve localization and
explanation. We adopt a similar decomposition principle, but for
zero-shot accident understanding: \textit{when} the impact occurs,
\textit{what} type of collision it is, and \textit{where} the impact is
localized.

\paragraph{Vision-language models and prompting.}
Recent VLM-based approaches use natural language prompts for traffic
scene reasoning, accident description, and safety-critical
understanding~\cite{zhang2025vlmaccident,shi2025scvlm,kim2025vruaccident}.
Concurrent work by Thakur and Talele~\cite{thakur2026modular} studies
zero-shot accident classification using multi-prompt textual
descriptions with CLIP-style retrieval. Our approach instead uses
structured prompts as complementary reasoning views for a generative
VLM, followed by vote aggregation and pairwise adjudication. This is
motivated by chain-of-thought prompting~\cite{wei2022cot},
self-consistency~\cite{wang2022selfconsistency}, Tree of
Thoughts~\cite{yao2023tot}, and Ask Me Anything~\cite{arora2022ama},
which show that intermediate reasoning and multiple prompt
formulations can improve robustness. We place this prompting strategy
inside a perception-first pipeline: frames are first temporally
selected, then classified, then spatially grounded.

%% file: sec/3_method.tex
\section{Methodology}
\label{sec:method}

\subsection{Temporal Detection}
\label{sec:temporal}

Accident detection in surveillance videos is sparse in time,
making most frames non-informative for a VLM.
We therefore first localize the relevant
temporal segment using Meta's Perception Encoder
(PE)~\cite{bolya2025pe}, a contrastive vision-language model similar
to CLIP~\cite{radford2021clip}. PE scores each sampled frame by its
semantic alignment with the text query \textbf{traffic accident},
allowing us to select the frames most likely to contain the impact.

Given a video $V$, we uniformly sample frames
$\{f_i\}_{i=1}^{N}$ at 8 FPS. Each frame $f_i$ is encoded into a
visual embedding $\mathbf{v}_i \in \mathbb{R}^d$, and the query text
\textbf{traffic accident} is encoded into $\mathbf{t}\in\mathbb{R}^d$.
We compute the cosine similarity

\begin{equation}
s_i = \frac{\mathbf{v}_i^\top \mathbf{t}}
{\|\mathbf{v}_i\| \, \|\mathbf{t}\|}.
\end{equation}

Frames are ranked by $s_i$ in descending order, with ties broken by
earlier timestamp, and the top-$K_{\text{PE}}$ frames are selected as
accident candidates. Let $\tau_j$ denote the timestamps
of these selected frames. We define the temporal window directly as
\[
\left[\min_j \tau_j - \delta,\; \max_j \tau_j + \delta\right],
\]
with $\delta=2$\,s. This produces a short accident-centric window
that preserves context around the impact while discarding most
irrelevant frames. The selected key frames $\mathcal{F}_{\text{key}}$
and the expanded window are then passed to the classification and
localization stages.
The final accident time is the midpoint of the expanded window,
\[
\hat{t}=\frac{(\min_j \tau_j-\delta)+(\max_j \tau_j+\delta)}{2}
=\frac{\min_j \tau_j+\max_j \tau_j}{2}.
\]
Thus only the earliest and latest selected peak timestamps affect
$\hat{t}$.

\subsection{Structured Multi-Prompt Classification}
\label{sec:classification}

After temporal localization, Stage~2 predicts the accident type from
the selected key frames $\mathcal{F}_{\text{key}}$. A single prompt is
often insufficient because accident categories depend on different cues
--- vehicle motion, contact geometry, impact angle, and whether one or
multiple vehicles are involved. We therefore query Qwen-3.5-VL~9B~\cite{bai2023qwenvl} with
five structured prompts over the same key frames and metadata context
$M$ (scene layout, weather, time of day, and video quality). The prompts
cover five complementary views: direct classification, temporal motion,
geometric contact reasoning, contrastive elimination, and tiebreaking.

Let $\mathcal{P}=\{p_1,\dots,p_5\}$ denote the prompt set. Each prompt
maps the key frames and metadata to a class label,
\begin{equation}
y_i = p_i(\mathcal{F}_{\text{key}}, M), \quad i \in \{1,\dots,5\}.
\end{equation}
The outputs are aggregated by the uncertainty-aware voting procedure
described below. Full prompt templates are provided in the supplementary
material.
\vspace{-5pt}
\paragraph{Aggregation and adjudication.}
Let $\mathcal{Y}=\{y_1,\dots,y_5\}$ denote the predictions of the
five prompts and $n_c=\sum_i \mathbb{1}[y_i=c]$ their vote counts.
We summarize the vote distribution by the top-two margin
$m = n_{(1)}-n_{(2)}$ and the normalized entropy
$\tilde{H} = -\tfrac{1}{\log_2 K}\sum_c \hat{p}_c \log_2 \hat{p}_c$
with $\hat{p}_c=n_c/|\mathcal{Y}|$, where $K$ is the number of distinct predicted classes.

If $m > \tau_m$ or $\tilde{H} \le \tau_H$
(default $\tau_m{=}2$, $\tau_H{=}0.75$), the plurality class
$\arg\max_c n_c$ is returned. Otherwise we escalate in two cheap
stages, both operating on a reduced subset of at most six evenly
spaced key frames:
(i) a structured \emph{tiebreaker} prompt that emits an additional
vote, after which the majority is re-taken; and, if the vote
remains ambiguous, (ii) a focused \emph{pairwise adjudication}
restricted to the top-two classes $(c_1,c_2)$, conditioned on
impact geometry and contact point. The pairwise output overrides
the plurality whenever it lies in $\{c_1,c_2\}$.

\subsection{Spatial Impact Localization}
\label{sec:spatial}

Given the key frames $\mathcal{F}_{\text{key}}$ from Stage~1 and
the predicted accident type $\hat{c}$ from Stage~2, this stage
localizes the impact in pixel coordinates using
OWL-v2~\cite{minderer2023owlv2}, an open-vocabulary detector.
\vspace{-5pt}
\paragraph{Type- and scene-conditioned prompting.}
Rather than querying the detector with a generic ``car crash'',
we construct a prompt set conditioned on both $\hat{c}$ and the
scene layout $\ell$ (e.g., \textit{highway}, \textit{signalized
intersection}) recovered from dataset metadata. Each base phrase
(e.g., ``car crashing into back of another car'' for
\textsc{rear-end}, ``side impact crash'' for \textsc{t-bone}) is
suffixed with a scene phrase such as ``on a highway''. This steers
the detector toward the contact region rather than every vehicle
in the scene.
\vspace{-5pt}
\paragraph{Multi-frame detection and aggregation.}
For each key frame, OWL-v2~\cite{minderer2023owlv2} returns detections
$\{(s_\ell, b_\ell)\}$ with confidence $s_\ell$ and box
$b_\ell = (x^{\min}_\ell, y^{\min}_\ell, x^{\max}_\ell, y^{\max}_\ell)$,
retained above a score threshold $\theta = 0.05$. Detections from
all key frames are pooled and the top-$K$ entries by score are
kept ($K{=}5$); denote this set $\mathcal{T}$.

The impact point $(\hat{x}, \hat{y})$ is the score-weighted
centroid of the top-$K$ box centers:
\begin{equation}
\hat{x} = \frac{\sum_{\ell \in \mathcal{T}} s_\ell\, c^{x}_\ell}{\sum_{\ell \in \mathcal{T}} s_\ell},
\quad
\hat{y} = \frac{\sum_{\ell \in \mathcal{T}} s_\ell\, c^{y}_\ell}{\sum_{\ell \in \mathcal{T}} s_\ell},
\end{equation}
where $c^{x}_\ell = (x^{\min}_\ell + x^{\max}_\ell)/2$ and
$c^{y}_\ell = (y^{\min}_\ell + y^{\max}_\ell)/2$. The impact
region is the axis-aligned union of the top-$K$ boxes, clamped to
the frame. This score-weighted aggregation implicitly enforces
temporal consistency: regions detected confidently across multiple
key frames dominate the centroid, while transient false positives
are downweighted.

%% file: sec/4_experiments.tex
\section{Experiments}
\label{sec:experiments}

\subsection{Dataset and Evaluation Protocol}
\label{sec:dataset}

We evaluate our method on the zero-shot ACCIDENT@CVPR~2026
benchmark~\cite{picek2026accident}. The benchmark provides no labeled
real-world training data. Instead, participants are given a synthetic
CARLA~\cite{deschaud2021kitti} development set with full annotations, including collision time,
impact coordinates, accident type, and per-frame bounding boxes. Final
evaluation is performed on real fixed-view CCTV clips.

The test videos are challenging due to low resolution, compression
artifacts, partial occlusions, and shallow camera angles.
For each clip, the system must predict
three outputs: accident time in seconds, impact location $(x,y)$ in
normalized frame coordinates, and accident type from
$\{$rear-end, T-bone, head-on, sideswipe, single$\}$. Coarse scene
metadata, such as layout, weather, and time of day, is provided with
the videos but is not used directly by the scorer.
\vspace{-8pt}
\paragraph{Evaluation Metric.}
The challenge reports three task scores in $[0,1]$. The temporal score
$\mathcal{T}$ and spatial score $\mathcal{S}$ are computed using
Gaussian-style similarities based on time error and Euclidean distance
to the ground-truth impact location, respectively. The classification
score $\mathcal{C}$ is top-1 accuracy. The final leaderboard score is
the harmonic mean:
\begin{equation}
\text{ACCIDENT score}
= \frac{3}{\frac{1}{\mathcal{T}} + \frac{1}{\mathcal{S}} + \frac{1}{\mathcal{C}}}
\end{equation}

While the official Kaggle leaderboard reports public and private overall scores, we report the per-component scores $\mathcal{T}$, 
$\mathcal{S}$, and $\mathcal{C}$, which we obtained from the updated leaderboard~\cite{picek2026accident}.
\subsection{Implementation Details}
\label{sec:impl}

All experiments are conducted on a single NVIDIA L4 GPU with 24\,GB of
memory. \textbf{Stage~1} uses the Perception Encoder~\cite{bolya2025pe}
(PE-Core-G14-448) in half precision to embed frames sampled at 8~FPS.
The top-5 similarity peaks define the candidate temporal window, which
is expanded by $\delta{=}2$\,s. \textbf{Stage~2} uses Qwen-3.5-VL~9B~\cite{bai2023qwenvl},
served locally via Ollama~\cite{ollama} (\texttt{qwen3.5vl:9b}) with 4-bit
quantisation, \texttt{num\_ctx}${=}12{,}288$, and temperature $0.2$.
Each clip is classified using up to 8 motion-scored key frames and 5
structured prompts. The adjudication step applies the entropy/margin
gate from Sec.~\ref{sec:classification}, with $\tau_m{=}2$ and
$\tau_H{=}0.75$, over at most 6 evenly spaced frames.
\textbf{Stage~3} uses \texttt{owlv2-base-patch16-ensemble}~\cite{minderer2023owlv2} with score
threshold $\theta{=}0.05$ and top-$K{=}5$ aggregation. 

\subsection{Main Results}
\label{sec:main-results}

Table~\ref{tab:main} reports the final ACCIDENT@CVPR~2026~\cite{picek2026accident} test score
of our pipeline, compared with two simple baselines that predict a fixed
time offset and the frame centre for every clip.

\begin{table}[t]
\centering
\caption{Results on the ACCIDENT@CVPR~2026~\cite{picek2026accident} test set.
The overall score is the harmonic mean of $\mathcal{T}$, $\mathcal{S}$, and $\mathcal{C}$.
\textsuperscript{$\dagger$}Rule-based, mid-clip time, frame center.
\textsuperscript{$\ddagger$}Rule-based, quarter-clip time, frame center.}
\label{tab:main}
\scriptsize
\setlength{\tabcolsep}{3pt}
\resizebox{\columnwidth}{!}{%
\begin{tabular}{l c c c c c}
\toprule
Method & Public LB & Private LB & $\mathcal{C}$ & $\mathcal{T}$ & $\mathcal{S}$ \\
\midrule
Baseline~A\textsuperscript{$\dagger$} & 0.2714 & 0.2734 & 0.5807 & 0.1896 & 0.2505 \\
Baseline~B\textsuperscript{$\ddagger$} & 0.3107 & 0.3188 & 0.5807 & 0.2664 & 0.2505 \\
\midrule
\textbf{Ours}                          & \textbf{0.3852} & \textbf{0.4015} & \textbf{0.5057} & \textbf{0.3689} & \textbf{0.3498} \\
\bottomrule
\end{tabular}%
}
\end{table}
\vspace{-4pt}

\subsection{Ablation Studies}
\label{sec:ablations}

We analyze the contribution of each stage in the pipeline by varying one component at a time, while maintaining the other two components at their full-pipeline settings. Since the leaderboard only provides the final harmonic mean score, all our analyses report this single metric.

\begin{table}[t]
\centering
\caption{Ablation studies. For each group, the other two stages are held
fixed at their full-pipeline settings.}
\label{tab:ablations}
\scriptsize
\setlength{\tabcolsep}{3pt}
\resizebox{\columnwidth}{!}{%
\begin{tabular}{llcc}
\toprule
Stage & Variant & Public & Private \\
\midrule
Temporal
  & Uniform midpoint & 0.3444 & 0.3592 \\
  & PE Top-1 frame & 0.3435 & 0.3627 \\
  & \textbf{PE $\delta$-window midpoint (Ours)} & \textbf{0.3852} & \textbf{0.4015} \\
\midrule
Classification
  & 1-prompt structured & 0.3801 & 0.3961 \\
  & 3-prompt majority vote & 0.3809 & 0.3978 \\
  & 5-prompt + tiebreaker & 0.3849 & 0.4001 \\
  & \textbf{+ entropy-gated pairwise (Ours)} & \textbf{0.3852} & \textbf{0.4015} \\
\midrule
Spatial
  & Molmo2 pointing~\cite{clark2026molmopoint} & 0.2589 & 0.2647 \\
  & Centre-of-frame $(0.5,\,0.5)$ & 0.3358 & 0.3487 \\
  & \textbf{OWL-v2~\cite{minderer2023owlv2} type+scene conditioned (Ours)} & \textbf{0.3852} & \textbf{0.4015} \\
\bottomrule
\end{tabular}%
}
\end{table}

The $\delta$-expanded temporal window improves the score by $0.039$
over PE Top-1, suggesting that the highest-similarity frame alone is a
noisy temporal estimate. Additional prompts and adjudication steps
provide consistent classification gains, improving the final score by
$0.0054$ over the single-prompt setting. OWL-v2~\cite{minderer2023owlv2} provides the largest stage-wise improvement, increasing the score by 0.053 over the center-of-frame baseline.

Across the ablation studies, the three stages contribute unevenly to the
final score. Spatial localisation provides the largest gain, followed by
temporal localisation, while the classification ensemble contributes a
smaller but consistent improvement. Most residual errors fall into
distant collisions where OWL-v2~\cite{minderer2023owlv2} either misses or selects a foreground
vehicle, and adverse capture conditions such as rain, night, or
occlusion.

\vspace{-5pt}

%% file: sec/5_conclusion.tex
\section{Conclusion}

\label{sec:conclusion}
In this paper, we presented a pipeline for zero-shot accident understanding in surveillance video that decomposes the task into \emph{when}, \emph{what}, and \emph{where}. The method combines temporal windowing, multi-prompt classification, and type-conditioned spatial grounding. On the ACCIDENT@CVPR 2026 benchmark, it outperforms a centre-of-frame baseline without any fine-tuning. Type-conditioned grounding in particular improves spatial localisation, indicating that grounding around the predicted accident type reduces foreground-object bias. Overall, task decomposition appears to be a useful design choice for zero-shot accident understanding in challenging surveillance footage.




%% file: sec/X_suppl.tex
\clearpage
\setcounter{page}{1}
\maketitlesupplementary

\tcbset{
  promptbox/.style={
    enhanced,
    breakable,
    boxrule=0.5pt,
    arc=2pt,
    left=4pt,
    right=4pt,
    top=4pt,
    bottom=4pt,
    fontupper=\small
  }
}

\section*{Overview}
This supplementary material includes:
\begin{itemize}
  \item Full text of the classification prompts, conditional tiebreaker,
    and pairwise adjudicator (Sec.~\ref{sec:supp-prompts}).
  \item Type- and scene-conditioned phrase templates used by the
    OWL-v2 spatial localizer (Sec.~\ref{sec:supp-owl-prompts}).
  \item Implementation hyperparameters and runtime details
    (Sec.~\ref{sec:supp-hyperparams}).
  \item Per-stage decision statistics
    (Sec.~\ref{sec:supp-decisions}).
  \item Qualitative failure modes
    (Sec.~\ref{sec:supp-qualitative}).
\end{itemize}

\section*{Classification Prompts}
\label{sec:supp-prompts}

All prompts share a common structured-output constraint and a common
metadata prefix. The metadata prefix is constructed per clip from the
four available metadata fields: scene layout, weather, time of day, and video quality.

\paragraph{Shared output constraint.}

\begin{tcolorbox}[promptbox,colback=gray!5,colframe=gray!50]
REQUIRED format --- follow this EXACTLY:\\
1. \textbf{Vehicles}: List every distinct vehicle you see. If only one
vehicle is visible, write ``Vehicle 1: [description]. No second vehicle
visible.''\\
2. \textbf{Contact}: Describe the contact point, or write
``None --- single vehicle incident'' if only one vehicle is visible.\\
3. \textbf{Answer}: \textless letter\textgreater
\end{tcolorbox}

\paragraph{$p_1$: Baseline.}

\begin{tcolorbox}[promptbox,colback=blue!3,colframe=blue!40]
You are given frames from a traffic accident video.\\
\{metadata\_context\}\\
Analyze the frames and classify the accident type.\\[2pt]
Classification categories:\\
\{categories\}\\
Rules:\\
$\bullet$ Classify based ONLY on vehicle direction and angle at the
moment of impact.\\
$\bullet$ Ignore motion before or after the collision.\\
$\bullet$ Use the metadata as supporting context, but rely primarily on
visual evidence.\\[2pt]
\{constraint\}
\end{tcolorbox}

\paragraph{$p_2$: Temporal motion.}

\begin{tcolorbox}[promptbox,colback=orange!3,colframe=orange!40]
You are given frames from a traffic accident video.\\
\{metadata\_context\}\\
Watch how vehicles move from the beginning to the moment of collision.
Pay attention to each vehicle's direction of travel at the exact moment
of impact.\\[2pt]
Classification categories:\\
\{categories\}\\
Rules:\\
$\bullet$ Track vehicle motion over time but classify based on the
collision moment only.\\
$\bullet$ Use the metadata as supporting context, but rely primarily on
visual evidence.\\[2pt]
\{constraint\}
\end{tcolorbox}

\paragraph{$p_3$: Contact geometry.}

\begin{tcolorbox}[promptbox,colback=green!3,colframe=green!40]
You are given frames from a traffic accident video.\\
\{metadata\_context\}\\
Focus on the spatial geometry at the moment of collision:\\
$\bullet$ What angle do the vehicles meet at?\\
$\bullet$ Which part of each vehicle makes contact?\\
$\bullet$ Are vehicles moving in the same, opposite, or perpendicular
directions?\\[2pt]
Classification categories:\\
\{categories\}\\
Rules:\\
$\bullet$ Use ONLY the angle and contact point at impact to classify.\\
$\bullet$ Use the metadata as supporting context, but rely primarily on
visual evidence.\\[2pt]
\{constraint\}
\end{tcolorbox}

\paragraph{$p_4$: Contrastive elimination.}

\begin{tcolorbox}[promptbox,colback=red!3,colframe=red!40]
You are given frames from a traffic accident video.\\
\{metadata\_context\}\\
Classify the accident by eliminating wrong categories:\\
$\bullet$ Same direction and one vehicle hits the other's rear $\to$ A\\
$\bullet$ Vehicles meet at $\sim 90^\circ$ and one hits the other's side $\to$ B\\
$\bullet$ Only one vehicle is involved $\to$ C\\
$\bullet$ Opposite directions and front-to-front collision $\to$ D\\
$\bullet$ Parallel side-to-side glancing contact $\to$ E\\[2pt]
Full categories:\\
\{categories\}\\
Rules:\\
$\bullet$ Classify based ONLY on the moment of impact.\\
$\bullet$ Use the metadata as supporting context, but rely primarily on
visual evidence.\\[2pt]
\{constraint\}
\end{tcolorbox}

\paragraph{$p_5$: Tiebreaker.}

This prompt is invoked only when the first four prompts produce a
high-uncertainty vote distribution.

\begin{tcolorbox}[promptbox,colback=purple!3,colframe=purple!40]
You are given a traffic accident video.\\
\{metadata\_context\}\\
Two expert classifiers disagree on this accident. Watch very carefully.
Focus on ONLY the moment of collision:\\
1. Count the vehicles involved.\\
2. Determine each vehicle's direction of travel at impact.\\
3. Identify the exact contact point.\\
4. Estimate the angle between the vehicles at impact.\\[2pt]
Classification categories:\\
\{categories\}\\
Rules:\\
$\bullet$ If only one vehicle is involved $\to$ C\\
$\bullet$ If two vehicles travel the same direction and one hits the rear $\to$ A\\
$\bullet$ If two vehicles meet at $\sim 90^\circ$ and one hits the side $\to$ B\\
$\bullet$ If two vehicles travel opposite directions, front-to-front $\to$ D\\
$\bullet$ If two vehicles are parallel with side-to-side glancing contact $\to$ E\\[2pt]
\{constraint\}
\end{tcolorbox}

\paragraph{Pairwise adjudicator (final stage).}

Invoked only if the vote remains ambiguous after the tiebreaker.

\begin{tcolorbox}[promptbox,colback=yellow!5,colframe=yellow!50!black]
You are given a traffic accident video.\\
\{metadata\_context\}\\
Final adjudication: choose ONLY between $c_1$ ($\text{name}_1$) and
$c_2$ ($\text{name}_2$). Decide using only impact-moment geometry and
contact point. Do not choose any other class.\\[2pt]
Allowed classes:\\
$\bullet$ $c_1$: $\text{name}_1$\\
$\bullet$ $c_2$: $\text{name}_2$\\[2pt]
\{constraint\}
\end{tcolorbox}

\section*{Spatial Localization Prompts}
\label{sec:supp-owl-prompts}

OWL-v2 receives phrases constructed by combining a type-conditioned base phrase with a scene-conditioned suffix. Table~\ref{tab:owl-prompts} lists the templates.

\begin{table}[t]
\centering
\caption{Type- and scene-conditioned prompt templates for OWL-v2.}
\label{tab:owl-prompts}
\small
\setlength{\tabcolsep}{4pt}
\begin{tabular}{p{0.38\linewidth}p{0.54\linewidth}}
\toprule
Predicted type $\hat{c}$ & Type-conditioned base phrase \\
\midrule
Rear-end & ``car crashing into back of another car'' \\
T-bone & ``side impact crash between two cars'' \\
Head-on & ``two cars colliding head-on'' \\
Sideswipe & ``two cars in side-to-side glancing contact'' \\
Single & ``single vehicle crash'' \\
\midrule
Scene layout $\ell$ & Scene suffix \\
\midrule
Highway & ``on a highway'' \\
Signalized intersection & ``at a signalized intersection'' \\
Simple intersection & ``at an intersection'' \\
Grade-separated intersection & ``on an overpass or interchange'' \\
City street & ``on a city street'' \\
Parking lot & ``in a parking lot'' \\
Tunnel & ``inside a tunnel'' \\
\bottomrule
\end{tabular}
\end{table}

\section*{Implementation Hyperparameters}
\label{sec:supp-hyperparams}

Table~\ref{tab:hyper} consolidates the numeric settings used across all three stages. All values were fixed on a small CARLA development subset and held constant for all test submissions.

\begin{table}[t]
\centering
\caption{Pipeline hyperparameters.}
\label{tab:hyper}
\small
\setlength{\tabcolsep}{3pt}
\begin{tabular}{p{0.18\linewidth}p{0.39\linewidth}p{0.33\linewidth}}
\toprule
Stage & Parameter & Value \\
\midrule
Stage 1 & Encoder & PE-Core-G14-448, fp16 \\
        & Frame sampling rate & 8 FPS \\
        & PE peaks $K_{\text{PE}}$ & 5 \\
        & Window expansion $\delta$ & 2\,s \\
\midrule
Stage 2 & Backbone & Qwen-3.5-VL 9B \\
        & Quantization & 4-bit, Ollama \\
        & Context window & 12{,}288 \\
        & Temperature & 0.2 \\
        & Max keyframes & 8, motion-scored \\
        & Adjudication frames & $\le 6$ \\
        & Margin threshold $\tau_m$ & 2 \\
        & Entropy threshold $\tau_H$ & 0.75 \\
\midrule
Stage 3 & Detector & OWL-v2 base patch16 ensemble \\
        & Score threshold $\theta$ & 0.05 \\
        & Top-$K$ aggregation & 5 \\
        & Aggregation rule & score-weighted centroid \\
\midrule
System  & GPU & 1$\times$ NVIDIA L4, 24\,GB \\
        & Fine-tuning & none \\
        & Proprietary APIs & none \\
\bottomrule
\end{tabular}
\end{table}

\section*{Decision-Stage Statistics}
\label{sec:supp-decisions}

Table~\ref{tab:decisions} reports the proportion of test clips resolved at each stage of the classification cascade. The entropy/margin gate dispatches the majority of clips with the cheap base ensemble, and the more expensive escalation stages are invoked only for ambiguous cases.

\begin{table}[t]
\centering
\caption{Decision-stage distribution on the ACCIDENT@CVPR 2026 test set ($n{=}2027$).}
\label{tab:decisions}
\small
\begin{tabular}{lcc}
\toprule
Stage & \# clips & Share \\
\midrule
Base, no escalation & 988 & 48.7\% \\
Tiebreaker $p_5$ & 305 & 15.0\% \\
Pairwise adjudication & 734 & 36.2\% \\
\midrule
Total & 2027 & 100\% \\
\bottomrule
\end{tabular}
\end{table}

\section*{Qualitative Failure Modes}
\label{sec:supp-qualitative}

Most residual errors fall into three main categories. First, distant
collisions on long-perspective cameras can cause OWL-v2 to return a
low-confidence detection on a foreground vehicle. Second, adverse
weather or lighting can degrade PE similarity and shift the selected
temporal window. Third, shallow-angle rear-end and sideswipe crashes can
remain ambiguous even after pairwise adjudication.

\section*{Reproducibility}
\label{sec:supp-repro}

All prompts, hyperparameters, and implementation details are provided
to ensure reproducibility. The code will be made publicly available on request. The system uses no fine-tuned weights or proprietary APIs,
and all components are based on publicly accessible models and tools.